\documentclass[11pt]{article}
\usepackage{amssymb,amsmath,amsthm,amsfonts}
\usepackage[backend=biber,style=numeric,sorting=nyt]{biblatex}
\usepackage{booktabs}
\usepackage{courier}
\usepackage{enumitem}
\usepackage{geometry}
\usepackage{graphicx}
\usepackage{listings}
\usepackage{multirow}
\usepackage[parfill]{parskip}
\usepackage{svg}
\usepackage{tikz}
\usepackage{xurl}

\usetikzlibrary{calc}
\usetikzlibrary{decorations.pathreplacing}
\usetikzlibrary{fit}
\usetikzlibrary{patterns.meta}
\usetikzlibrary{shapes.geometric}

\usepackage{color}
\definecolor{deepblue}{rgb}{0,0,0.5}
\definecolor{deepred}{rgb}{0.6,0,0}
\definecolor{deepgreen}{rgb}{0,0.5,0}

\title{Inferno: An Extensible Framework for Spiking Neural Networks}
\author{Marissa Dominijanni\\University at Buffalo, Department of Computer Science and Engineering\\\texttt{mdomini@buffalo.edu}}

\date{2024 September 17}

\begin{document}

\maketitle

\begin{abstract}
    \noindent
    This paper introduces Inferno, a software library built on top of PyTorch that is designed to meet distinctive challenges of using spiking neural networks (SNNs) for machine learning tasks. We describe the architecture of Inferno and key differentiators that make it uniquely well-suited to these tasks. We show how Inferno supports trainable heterogeneous delays on both CPUs and GPUs, and how Inferno enables a ``write once, apply everywhere'' development methodology for novel models and techniques. We compare Inferno's performance to BindsNET, a library aimed at machine learning with SNNs, and Brian2/Brian2CUDA which is popular in neuroscience. Among several examples, we show how the design decisions made by Inferno facilitate easily implementing the new methods of Nadafian and Ganjtabesh in delay learning with spike-timing dependent plasticity.
\end{abstract}

\section{Introduction}
Spiking neural networks (SNNs) have lagged other neural network models, especially among those used by the general public. The deep learning community has largely coalesced around three major frameworks: PyTorch \cite{paszke_pytorch_2019}, TensorFlow \cite{abadi_tensorflow_2016}, and JAX \cite{frostig_compiling_2018}, plus HuggingFace for transformer models \cite{wolf_transformers_2020,vaswani_attention_2017}. Brian 2 \cite{stimberg_brian_2019} and its GPU-accelerated variants Brian2GeNN \cite{stimberg_brian2genn_2020} and Brian2CUDA \cite{alevi_brian2cuda_2022} are frequently used in the neuroscience community, but their focus on low-level detail and lack of interoperability with any of the major deep-learning frameworks limit their use in the broader context of machine learning. BindsNET \cite{hazan_bindsnet_2018} attempts to solve these issues, using PyTorch for high-performance tensor operations and focusing on a higher-level machine-learning oriented design, but sacrifices extensibility and the simulation of certain properties of SNNs to do so. We introduce Inferno, an SNN framework integrated with PyTorch, designed for higher-level machine-learning applications, and developed with the generalizability of components and extensibility thereof in mind.

The incorporation of \emph{time} is distinctive to SNNs, whose dynamics are described in the language of differential equations \cite{yamazaki_spiking_2022}. This should confer benefits, enough that SNNs have been called the ``third generation of neural network models'' \cite{maass_networks_1997}. Yet artificial neural networks (ANNs) continue to dominate---even in areas such as sequence modeling where the temporality of SNNs should excel \cite{brown_language_2020,openai_gpt-4_2023,jiang_mixtral_2024}. Prominent among the vicissitudes of temporality that retard adoption is the non-differentiability of spike trains, which makes training via back-propagation cumbersome \cite{saiin_spike_2024}. In spite of these challenges, back-propagation based methods \cite{saiin_spike_2024,lee_training_2016,lee_enabling_2020} as well as biologically plausible plasticity rules \cite{florian_reinforcement_2007,diehl_unsupervised_2015} have been successfully employed for training SNNs.

We regard the current dearth of SNNs among state-of-the-art deep-learning models as primarily a matter of infrastructure. Software libraries for the simulation of SNNs are in no short supply, but those oriented towards machine learning are. We developed Inferno in response to the scarcity of tools and the defects in those that exist, in order to bridge the divide between SNNs and the broader world of machine learning. Our principal design objectives were to develop a framework with support for simulating large networks of spiking neurons, training these networks with biologically plausible plasticity rules, while enabling the rapid extension of this framework to facilitate the development of new methods in the realm of SNNs at the pace demanded by contemporary deep-learning research. We prioritized developing a framework with a strong separation of concerns to achieve this, as well as incorporating native support for connection delays---an active and important area of SNN research \cite{iakymchuk_simplified_2015,madadi_asl_dendritic_2017,madadi_asl_dendritic_2018}. In this paper, we give a comprehensive overview of the Inferno framework, explain our approach to implementing these objectives, and offer a comparative evaluation of Inferno with other frameworks for simulating SNNs---especially within the realm of machine learning.

\section{Architecture}
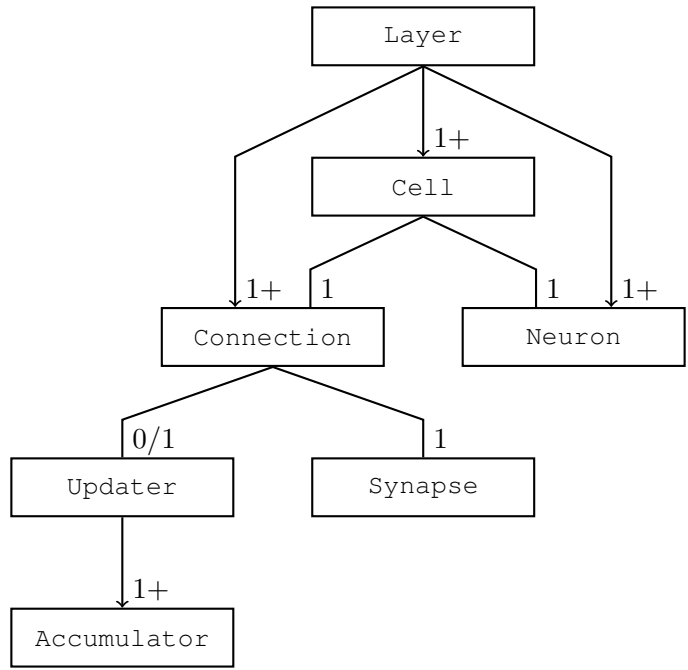
\begin{figure}[ht]
    \centering
    \begin{tikzpicture}[
        component/.style ={rectangle, draw=black, thick, fill=white, text width=7em, text centered, minimum height=2em},
        arity/.style ={rectangle, draw=none, fill=none, align=left, text height=1em, text depth=0.1em},
    ]
        \node[component] (layer) at (0,0) {\lstinline|Layer|};
        \node[component] (cell) at (0,-2) {\lstinline|Cell|};
        \node[component] (conn) at (-2,-4) {\lstinline|Connection|};
        \node[component] (neur) at (2,-4) {\lstinline|Neuron|};
        \node[component] (syn) at (0,-6) {\lstinline|Synapse|};
        \node[component] (updater) at (-4,-6) {\lstinline|Updater|};
        \node[component] (acc) at (-4,-8) {\lstinline|Accumulator|};

        \node[arity, anchor=south west] (lcell) at ($(cell.north) - (0, 0.05)$) {1+};
        \node[arity, anchor=south west] (lc) at ($(conn.north) - (0.5, 0) - (0, 0.05)$) {1+};
        \node[arity, anchor=south west] (cc) at ($(conn.north) + (0.5, 0) - (0, 0.05)$) {1};
        \node[arity, anchor=south west] (ln) at ($(neur.north) + (0.5, 0) - (0, 0.05)$) {1+};
        \node[arity, anchor=south west] (cn) at ($(neur.north) - (0.5, 0) - (0, 0.05)$) {1};
        \node[arity, anchor=south west] (cu) at ($(updater.north) - (0, 0.05)$) {0/1};
        \node[arity, anchor=south west] (cs) at ($(syn.north) - (0, 0.05)$) {1};
        \node[arity, anchor=south west] (ua) at ($(acc.north) - (0, 0.05)$) {1+};

        \draw[->, thick] (layer) edge[thick] (cell);
        \draw[->, thick] (layer.south) -- ($(-2.5, 0) + (cell.north)$) -- ($(conn.north) - (0.5, 0)$);
        \draw[->, thick] (layer.south) -- ($(2.5, 0) + (cell.north)$) -- ($(neur.north) + (0.5, 0)$);
        \draw[thick] (cell.south) -- ($(conn.north) + (0.5, 0.5)$) -- ($(conn.north) + (0.5, 0)$);
        \draw[thick] (cell.south) -- ($(neur.north) + (-0.5, 0.5)$) -- ($(neur.north) - (0.5, 0)$);
        \draw[thick] (conn.south) -- ($(updater.north) + (0, 0.5)$) -- (updater.north);
        \draw[thick] (conn.south) -- ($(syn.north) + (0, 0.5)$) -- (syn.north);
        \draw[->, thick] (updater.south) -- (acc.north);
    \end{tikzpicture}
    \caption{Entity relationship diagram showing object composition among the major components for modeling SNNs using Inferno. Numerical labels indicate the number of composed objects, where ``0/1'' indicates ``zero or one'' and ``1+'' indicates ``one or more''.}
    \label{fig:inferno-arch}
\end{figure}

In PyTorch, the various components of ANNs are modeled through subclasses of \lstinline|nn.Module|. Inferno continues this paradigm by having SNN components be modeled by extensions of the subclass \lstinline|Module|, which itself extends \lstinline|nn.Module|. The manner in which these components are composed with each other is shown in figure \ref{fig:inferno-arch}.

\subsection{Base Neural Components}
The most fundamental requirement for an SNN simulation library is to simulate the basic elements of a spiking network. What constitutes the ``basic elements'' however is a design decision that must be made. Elements in those libraries designed for the neuroscience community, such as in Brian 2 and BrainPy, typically take a numerical integration approach. There the user describes their model as a differential equation, the latter of which has ion channels as the most basic unit of its simulation \cite{stimberg_brian_2019,wang_brainpy_2023}. Those that instead target the machine learning community, such as BindsNET and PySNN\footnote{PySNN appears be no longer under active development. At the time of writing, the most recent commit in its GitHub repo was made on November $\text{28}^\text{th}$ 2019.}, treat a neuron as an indivisible base unit \cite{hazan_bindsnet_2018, buller_basbullerpysnn_2024}. Rather than using a numerical integration approach to implement either the solution of the model's representative differential equations or an approximation thereof, they implement the model directly using PyTorch primitives, as an approximation via Euler's method if no analytic solution exists. Inferno takes much of the latter direct approach but deviates in ways that broaden its simulation capabilities and simplify user extensions.

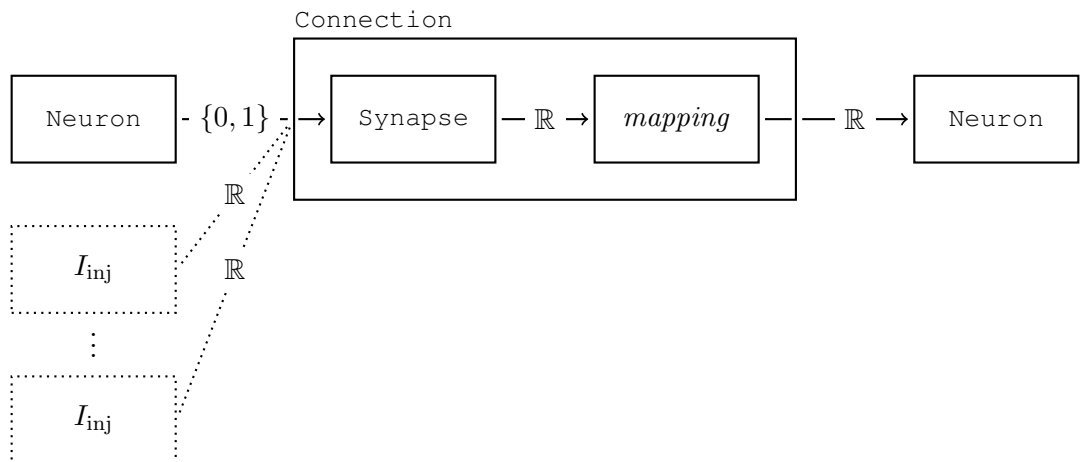
\begin{figure}[ht]
    \centering
    \begin{tikzpicture}[
        component/.style ={rectangle, draw=black, thick, fill=white, text width=5em, text centered, minimum height=3em},
        optional/.style ={rectangle, draw=black, thick, dotted, fill=white, text width=5em, text centered, minimum height=3em},
    ]
        \node[component] (nin) at (0,0) {\lstinline|Neuron|};
        \node[component] (nou) at (12,0) {\lstinline|Neuron|};
        \node[component] (syn) at (4.25,0) {\lstinline|Synapse|};
        \node[component] (map) at (7.75,0) {\emph{mapping}};
        \node[rectangle, draw=black, thick, fit=(syn) (map), inner sep=1.25em, label=above left:{\lstinline|Connection|}] (conn) {};
        \node[optional] (ixi) at (0,-2) {$I_\text{inj}$};
        \node[optional] (ixf) at (0,-4) {$I_\text{inj}$};
        \node[text centered] (ixdot) at (0, -2.9) {$\vdots$};
        \draw[shorten >= 2pt, shorten <= 2pt] (nin) edge[thick] node[midway, fill=white] {$\{0,1\}$} (conn);
        \draw[<-, shorten >= 2pt, shorten <= 2pt] (syn) edge[thick] (conn);
        \draw[->, shorten >= 2pt, shorten <= 2pt] (syn) edge[thick] node[midway, fill=white] {$\mathbb{R}$} (map);
        \draw[shorten >= 2pt, shorten <= 2pt] (map) edge[thick] (conn);
        \draw[->, shorten >= 2pt, shorten <= 2pt] (conn) edge[thick] node[midway, fill=white] {$\mathbb{R}$} (nou);
        \draw[shorten >= 3pt, shorten <= 3pt] (ixi.east) edge[thick, dotted] node[midway, fill=white] {$\mathbb{R}$} (conn.west);
        \draw[shorten >= 3pt, shorten <= 3pt] (ixf.east) edge[thick, dotted] node[midway, fill=white] {$\mathbb{R}$} (conn.west);
    \end{tikzpicture}
    \caption{Representation of the flow of data (tensors) between different kinds of neural components. The ``[\emph{mapping}]'' represents the internal logic unique to that \lstinline|Connection| (such as matrix multiplication with the learned weights). The $I_\text{inj}$ terms are optional and not supported by every \lstinline|Synapse|, but represent current injected into the synapse. The classes \lstinline|Neuron| and \lstinline|Synapse| have singular names because each encapsulates a single tensor per state parameter. Each object may represent an arbitrary number of simulated neurons and synapses respectively.}
    \label{fig:inferno-dataflow}
\end{figure}

\subsubsection{Neurons} \label{sec:inferno-neurons}
Inferno's \lstinline|Neuron| abstract class is used to represent a group of neurons with a common mode of operation (the model of neuronal dynamics and the hyperparameters that control the behavior of said model). As shown in figure \ref{fig:inferno-dataflow}, each \lstinline|Neuron| takes a single tensor of input currents and returns a tensor of spikes. The neuronal dynamics are implemented as a mixture of exact solutions where feasible and as approximations using Euler's method where not.

While this approach mirrors that of the machine-learning oriented frameworks, it departs in the specifics of its implementations. Each \lstinline|Neuron| is a single component represented as a \lstinline|Module|, as is the case with BindsNET and PySNN---both also based on PyTorch. Inferno, however, divides the behavior of a neuron up into reusable functional parts provided in the \lstinline|inferno.neural.functional| submodule, so as to facilitate extensions.

To illustrate, consider three neuronal models: leaky integrate-and-fire (LIF) \cite{burkitt_review_2006}, adaptive leaky integrate-and-fire (ALIF) \cite{bellec_long_2018}, and leaky integrate-and-fire with biologically defined reset rules ($\text{GLIF}_\text{2}$) \cite{teeter_generalized_2018}. All three of these models integrate external current as a function of membrane voltage, and define how this voltage decays back to a resting value according to the same linear dynamics. This process is implemented by the \lstinline|voltage_integration_linear| function. This is modeled as:

\begin{align*}
    \tau_m \frac{dV_m(t)}{dt} &= V_\text{rest} - V_m(t) + R_m I(t) &\text{when } V_m(t) < \Theta(t), \\
    V_m(t) &= V_\text{reset}(t) &\text{when } V_m(t) \geq \Theta(t),
\end{align*}

where $V_m(t)$ is the membrane voltage at time $t$, $\tau_m$ is the time membrane time constant, $V_\text{rest}(t)$ is the resting membrane voltage, $V_\text{reset}$ is the voltage to which the membrane is reset, $R_m$ is the membrane resistance, $\Theta(t)$ is the firing threshold, and $I(t)$ is the external current applied.

Both ALIF and $\text{GLIF}_\text{2}$ include an adaptive threshold that is incremented by a constant when an action potential is generated, and decays towards zero otherwise. The adaptive component of this threshold can be calculated using \lstinline|adaptive_thresholds_linear_spike| and applied with \lstinline|apply_adaptive_thresholds|. This is in contrast to LIF, where $\Theta(t)$ is invariant over time.

$\text{GLIF}_\text{2}$ changes how the voltage to which a neuron's membrane is set after generating an action potential is calculated. Rather than being set to a constant given as a hyperparameter, it is made to vary linearly with the difference between membrane voltage when an action potential is generated and the neuron's resting potential. The \lstinline|voltage_thresholding_linear| function can be used to calculate and apply this. Whereas in LIF and ALIF neurons $V_\text{reset}(t)$ is invariant over time. This constant value can be applied using \lstinline|voltage_thresholding_constant|.

Inferno uses this pattern internally in the above cases as well as others, such as extending quadratic integrate-and-fire (QIF) neurons to Izhikevich neurons \cite{izhikevich_dynamical_2007}, or exponential integrate-and-fire (EIF) neurons to adaptive exponential integrate-and-fire (AdEx) neurons \cite{brette_adaptive_2005}. In both of these cases, adaptive inhibitory input currents are added with \lstinline|adaptive_currents_linear| and applied with \lstinline|apply_adaptive_currents|.

This design exposes the relatedness between various neuronal models and allows for more flexibility than only providing complete implementations of neurons to the end user (although complete implementations are of course provided). This makes it far more approachable to the general machine learning community than requiring them to describe a neuron model with one or more differential equations.

\subsubsection{Synapses} \label{sec:inferno-synapses}
Unlike BindsNET and PySNN, Inferno explicitly models synapses. It does so with the \lstinline|Synapse| abstract class, which is used to represent a group of synapses with a common mode of operation (the model of synaptic kinetics and the hyperparameters which control the behavior of said model). As shown in figure \ref{fig:inferno-dataflow}, each \lstinline|Synapse| takes one or more input tensors, generally representing input spikes and injected current (although these are allowed to vary between models), and outputs the synaptic currents.

To show an example where the number of inputs may vary, we will consider the two synapse models Inferno includes out of the box: \lstinline|DeltaCurrentSynapse|, which implements the delta synapse model, and \lstinline|DeltaPlusCurrentSynapse| which implements the same model but allows for injected current \cite{rosenbaum_modeling_2024}. The former uses only a single input (a tensor of spikes). The latter accepts any number of tensors, the first of which is the tensor of spikes. Any subsequent tensors are treated as injected current. Inferno differentiates these two because, in the former, synaptic currents can be derived from spikes and therefore do not need to be stored separately.

Inferno leverages the \lstinline|Synapse| class in order to support delayed connections while maintaining a separation of concerns. Each \lstinline|Synapse| stores a rolling window of its past states (the synaptic current and input spikes) which can be accessed by a \lstinline|Connection| based on its learned delays. The mechanics of this storage will be discussed in section \ref{sec:observation-records}. This design requires a \lstinline|Synapse| to be constructed with some arguments dependent upon the \lstinline|Connection| with which it is associated. Inferno achieves this using a dependency injection approach. Every \lstinline|Synapse| has a method \lstinline|partialconstructor| which accepts the arguments not dependent on the \lstinline|Connection| and returns a constructor function. The constructor's signature is specified by the \lstinline|SynapseConstructor| protocol, which takes the shape of the inputs, the length of the simulation step, the maximum supported delay, and the batch size.

\subsubsection{Connections} \label{sec:inferno-connections}
Inferno's \lstinline|Connection| abstract class is used to represent different configurations of trainable mappings between two groups of neurons. Each \lstinline|Connection| owns a \lstinline|Synapse|, constructed according to a partial function of its constructor as described in section \ref{sec:inferno-synapses}, which it uses for delays and for converting the discrete input spikes into real values representing electrical current. Each \lstinline|Connection| takes inputs like its owned \lstinline|Synapse| (although with a potentially different shape, as defined by the \lstinline|Connection| itself).

Unlike in BindsNET or PySNN, connections in Inferno support trainable delays, and are designed to interface with their owned \lstinline|Synapse| such that implementing a new type of connection does not require re-inventing delay support. Each \lstinline|Connection| is required to implement a property \lstinline|selector| which returns a tensor of delays the \lstinline|Synapse| uses to select the delay-adjusted synapse state. This is discussed in-detail in section \ref{sec:observation-records}.

\lstinline|Connection| objects in Inferno also implement an interface to facilitate generalizable updates, such that each plasticity-based update rule does not need to be implemented for each connection type, as is the case with BindsNET. This is discussed in detail in section \ref{sec:generalized-updates}, and is facilitated using the generalized presynaptic and postsynaptic reshaping described in section \ref{sec:generalized-shapes}.

\subsection{Networks, Monitoring, and Training}
While \lstinline|Neuron|, \lstinline|Synapse|, and \lstinline|Connection| provide the essential building-blocks of SNNs, they alone do not suffice when it comes to implementing and training more complex models. To keep with Inferno's design philosophy of maximizing generalizability, and to facilitate the rapid development and prototyping of biologically-plausible local learning methods, we introduce additional structures that enable a ``don't repeat yourself'' approach to implementing such methods.

\subsubsection{Layers and Cells} \label{sec:inferno-layers-cells}
In the context of ANNs, the convention is to refer to the mapping from inputs to outputs as a \emph{layer}. This includes both the trainable mapping (such as a linear or convolutional connection) and the nonlinearity (such as a rectifier or logistic sigmoid). The natural equivalent of this in the modeling of SNNs is the trainable mapping (a \lstinline|Connection|) and the neurons which receive input from that connection (a \lstinline|Neuron|). In the parlance of Inferno, this is a \lstinline|Cell|.

\lstinline|Cell| objects are not used for inference, but are instead a construct for simplifying the implementation of biologically-plausible learning methods. In frameworks such as BindsNET, which is arguably the current SNN framework best-suited to general machine-learning tasks, each training method must be implemented for each type of connection. Such a process which development and limits the scope to which a newly developed method can be applied by other researchers and developers.

On their own, \lstinline|Cell| objects cannot provide fully generalizable and efficient access for training. \lstinline|Layer| objects enable support for inference on \lstinline|Cell| objects as well as training their parameters. Each \lstinline|Layer| associates one or more \lstinline|Connection| objects with one or more \lstinline|Neuron| objects. How the outputs of connections are mapped to the inputs of neurons is fully customizable by subclassing \lstinline|Layer| and defining the \lstinline|wiring| method for it. Out of the box, Inferno includes \lstinline|Biclique| and \lstinline|Serial| as layer architectures. Both define an all-to-all relationship between connections and neurons. The former supports an arbitrary number of either, while the latter only supports one of each but strips away the indirection required for situations where a more flexible architecture is needed.

Inferno makes extensive use of PyTorch's system of \emph{hooks} (specifically \emph{forward hooks}), which are called when data is passed forward through a \lstinline|Module|. With Inferno's \lstinline|observe| submodule, the states of neural components can be tracked and recorded with \lstinline|Monitor| objects and the state simplified with \lstinline|Reducer| objects. With \lstinline|Cell| and \lstinline|Layer| objects, the collection of these values can be redirected to the call of the \lstinline|Layer| itself so a single \lstinline|Connection| or \lstinline|Neuron| can be used across layers if necessary. To handle cases where complex, multistep calculations need to occur on intermediate states, \lstinline|Cell| implements \lstinline|Observable|, so monitors can monitor other monitors.

\subsubsection{Example: Implementing New Layer Architectures} \label{sec:inferno-new-layers}
As previously stated, the \lstinline|Layer| object supports a flexible architecture. Here we will consider a practical example of a layer that implements the model described by Diehl and Cook \cite{diehl_unsupervised_2015}.

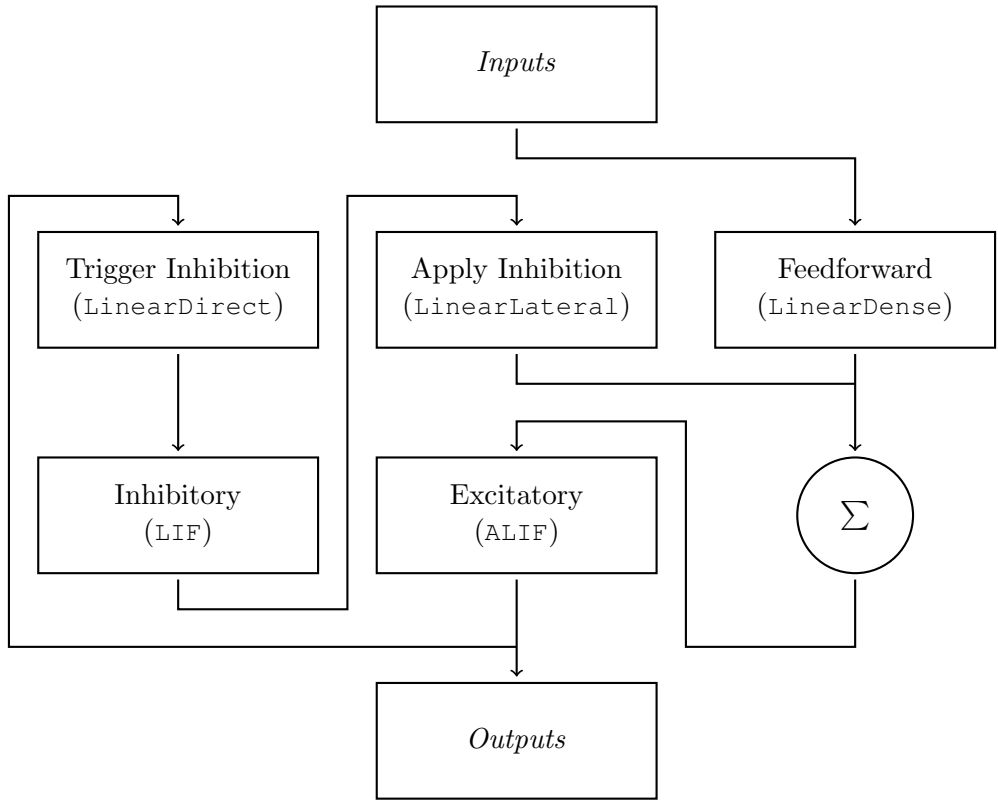
\begin{figure}[ht]
    \centering
    \begin{tikzpicture}[
        component/.style ={rectangle, draw=black, thick, fill=white, text width=9em, text centered, minimum height=4em},
        thincomponent/.style ={rectangle, draw=black, thick, fill=white, text width=9em, text centered, minimum height=2em},
        op/.style ={circle, draw=black, thick, fill=white, text width=2em, text centered, minimum height=4em},
    ]
        \node[component] (input) at (0,3) {\emph{Inputs}};
        \node[component] (exc2inh) at (-4.5,0) {Trigger Inhibition\\(\lstinline|LinearDirect|)};
        \node[component] (inh2exc) at (0,0) {Apply Inhibition\\(\lstinline|LinearLateral|)};
        \node[component] (ff) at (4.5,0) {Feedforward\\(\lstinline|LinearDense|)};
        \node[component] (inh) at (-4.5,-3) {Inhibitory\\(\lstinline|LIF|)};
        \node[component] (exc) at (0,-3) {Excitatory\\(\lstinline|ALIF|)};
        \node[op] (sum) at (4.5,-3) {$\sum$};
        \node[component] (output) at (0,-6) {\emph{Outputs}};

        \draw[->, thick, shorten >= 2pt, shorten <= 2pt] (input.south) -- (0, 1.75) -- (4.5, 1.75) -- (ff.north);
        \draw[->, thick, shorten >= 2pt, shorten <= 2pt] (ff.south) -- (sum.north);
        \draw[thick, shorten <= 2pt] (inh2exc.south) |- (4.5, -1.25);
        \draw[->, thick, shorten >= 2pt, shorten <= 2pt] (sum.south) -- (4.5, -4.75) -- (2.25, -4.75) -- (2.25, -1.75) -- (0, -1.75) -- (exc.north);
        \draw[->, thick, shorten >= 2pt] (0, -4.75) -- (-6.75, -4.75) -- (-6.75, 1.25) -- (-4.5, 1.25) -- (exc2inh.north);
        \draw[->, thick, shorten >= 2pt, shorten <= 2pt] (exc.south) -- (output.north);
        \draw[->, thick, shorten >= 2pt, shorten <= 2pt] (inh.south) -- (-4.5, -4.25) -- (-2.25, -4.25) -- (-2.25, 1.25) -- (0, 1.25) -- (inh2exc.north);
        \draw[->, thick, shorten >= 2pt, shorten <= 2pt] (exc2inh.south) -- (inh.north);
    \end{tikzpicture}
    \caption{Diagram of the model proposed by Diehl and Cook for classification on the MNIST dataset \cite{diehl_unsupervised_2015}. The text shown in parenthesized monospace shows the Inferno class corresponding to the model component in the original work.}
    \label{fig:dc2015-arch}
\end{figure}

This architecture includes two groups of neurons---one excitatory and one inhibitory---as well as three connections. One of these connections is responsible for incorporating the external input (which is gotten by encoding the MNIST images in \cite{diehl_unsupervised_2015}), while the other two are responsible for managing the lateral inhibition. One applies the inhibitory signal stimulated by the excitatory output, and the other excites the neurons that generate the inhibitory signal. Below we will include a simple implementation of this using the \lstinline|Layer| class.

\begin{lstlisting}[caption={Simple implementation of an excitatory-inhibitory layer.}, label=lst:impl-dc2015, language=Python]
from inferno.neural import Layer, Connection, Neuron
import torch

class ExcInhLayer(Layer):
    def __init__(
        self,
        fwd: Connection,
        exc2inh: Connection,
        inh2exc: Connection,
        exc: Neuron,
        inh: Neuron,
    ):
        Layer.__init__(self)

        self.add_connection("fwd", fwd)
        self.add_connection("exc2inh", exc2inh)
        self.add_connection("inh2exc", inh2exc)
        self.add_neuron("exc", exc)
        self.add_neuron("inh", inh)
        self.add_cell("fwd", "exc")

    def wiring(
        self, inputs: dict[str, torch.Tensor], **kwargs
    ) -> dict[str, torch.Tensor]:
        return {
            "exc": inputs["fwd"] + inputs["inh2exc"],
            "inh": inputs["exc2inh"]
        }
\end{lstlisting}

When implementing this at the library-level, we would want to add some additional functionality in the form of buffers for the recursive connections to prevent end-users from needing to pass them in manually, which would include a wrapper around the \lstinline|forward| method. At the application level however, this strategy enables an end-user to implement a generalized template for a complex model that includes the ability to experiment with different components beyond those considered in \cite{diehl_unsupervised_2015}. For example, inhibition models other than lateral inhibition can be experimented with by using different types of connections.

\subsubsection{Training} \label{sec:inferno-training}
Inferno facilitates the updating of model parameters through the use of trainers. At this stage, Inferno includes \lstinline|CellTrainer|, atop which various local training paradigms can be implemented. \lstinline|CellTrainer| provided a simplified interface for adding and removing \lstinline|Cell| objects from the trainer (used when training a model), as well as adding and removing monitors from them when added or removed from the trainer (used when implementing a new trainer). It utilizes \lstinline|MonitorPool| to share monitors between components when it is safe to do so. For example, if a \lstinline|Neuron| takes as input the merged output from two or more \lstinline|Connection| objects, so long as both corresponding \lstinline|Cell| objects are registered to the same trainer, monitors for attributes such as the postsynaptic spikes will not be duplicated. Currently, \lstinline|IndependentCellTrainer| provides an additional layer of abstraction for cases where the training of each \lstinline|Cell| is independent of any others, such as for spike-timing-dependent plasticity (STDP) methods.

\section{Critical Features}
\subsection{Recording and Accessing Observations Over Time} \label{sec:observation-records}
The simulation of propagation delays in spiking neural networks is an area of active research \cite{madadi_asl_dendritic_2018,xiao_temporal_2024,deckers_co-learning_2024} and supporting this was a primary design goal of Inferno. These delays represent the amount of time it takes until an action potential is integrated into subsequent neurons. While Brian 2 supports heterogeneous delays \cite{stimberg_brian_2019}, support is much more limited across frameworks with GPU acceleration. BindsNET has no support for any synaptic delays \cite{hazan_bindsnet_2018}, while Brian2GeNN only supports homogeneous delays \cite{stimberg_brian2genn_2020}, Brian2CUDA does include support for GPU-accelerated heterogeneous delays \cite{alevi_brian2cuda_2022}. Inferno supports trainable GPU-accelerated heterogeneous by providing an interface which leverages gather and scatter operations as described in \cite{he_efficient_2007} as well as tensor slicing and advanced tensor indexing to perform these operations efficiently \cite{paszke_pytorch_2019}.

The most general interface for this support provided by Inferno is the \lstinline|RecordTensor| class, which uses a reflective programming approach in combination with \lstinline|Module| (the extension of PyTorch's \lstinline|nn.Module| used throughout Inferno). A \lstinline|RecordTensor| stores a rolling window of observations (spikes, synaptic currents, etc.) along with a fixed step time that is assumed to have elapsed between each observation. See figure \ref{fig:rt-storage}.

\begin{figure}[ht]
    \centering
    \begin{tikzpicture}[
        slice/.style ={rectangle, draw=black, fill=white, text width=9em, text centered, minimum height=2em, rotate=90},
        hiddenslice/.style ={rectangle, draw=none, fill=none, text centered, minimum size=2em},
    ]
        \node[slice] (si) at (-3,0) {};
        \node[hiddenslice] (sci) at (-2,0) {$\cdots$};
        \node[slice] (sdt) at (-1,0) {$\Delta t$};
        \node[slice] (s0) at (0,0) {$0$};
        \node[slice] (st) at (1,0) {$T$};
        \node[slice] (stdt) at (2,0) {$T - \Delta t$};
        \node[hiddenslice] (scf) at (3,0) {$\cdots$};
        \node[slice] (sf) at (4,0) {};

        \draw[decorate, decoration={brace,amplitude=0.5em,raise=0.5em}] (si.north east) -- (sf.south east) node[midway, above=1.5em] {$\lfloor T / \Delta t \rfloor + 1$};

        \draw[decorate, decoration={brace,amplitude=0.5em,raise=0.5em}] (si.north west) -- (si.north east) node[midway, rotate=90, above=1.5em] {$N_1 \times \cdots \times N_d$};
    \end{tikzpicture}
    \caption{Each of the $\lfloor T / \Delta t \rfloor + 1$ time slices are stored contiguously in memory, and a pointer $p$ to the index at which $t = T$ is tracked and represents the next location to write to.}
    \label{fig:rt-storage}
\end{figure}
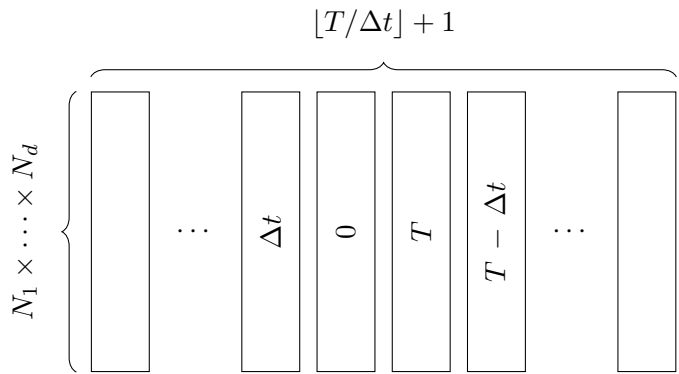

\begin{figure}[ht]
    \centering
    \begin{tikzpicture}[
        ]
            \matrix [nodes={draw=none, fill=gray!20, minimum size=3em},column sep={0em,between borders}, anchor=east] (record) at (-2,0)
            {
                \node {$a_0$}; & \node {$b_0$}; & \node[fill=purple!30] {$c_0$}; & \node[fill=violet!30] {$d_0$}; \\
                \node[fill=blue!30] {$a_1$}; & \node {$b_1$}; & \node[fill=purple!30] {$c_1$}; & \node {$d_1$}; \\
                \node[fill=blue!30] {$a_2$}; & \node[fill=teal!30] {$b_2$}; & \node[fill=green!30] {$c_2$}; & \node {$d_2$}; \\
                \node {$a_3$}; & \node[fill=red!30, postaction={pattern={Lines[angle=45, distance=0.5em, line width=0.25em]}, pattern color=teal!30}] {$b_3$}; & \node[fill=green!30] {$c_3$}; & \node {$d_3$}; \\
                \node[fill=orange!30] {$a_4$}; & \node {$b_4$}; & \node {$c_4$}; & \node {$d_4$}; \\
                \node[fill=orange!30] {$a_5$}; & \node {$b_5$}; & \node {$c_5$}; & \node[fill=yellow!30] {$d_5$}; \\
            };
            \node at ($(record) - (0,4)$) {Record};
            \matrix [nodes={draw=none, minimum size=3em},column sep={0em,between borders}] (selects) at (0,0)
            {
                \node[fill=blue!30] {$1.5$}; & \node[fill=orange!30] {$4.9$}; \\
                \node[fill=red!30] {$3$}; & \node[fill=teal!30] {$2.1$}; \\
                \node[fill=green!30] {$2.75$}; & \node[fill=purple!30] {$0.3$}; \\
                \node[fill=yellow!30] {$5$}; & \node[fill=violet!30] {$0$}; \\
            };
            \node at ($(selects) - (0,4)$) {Selector};
            \matrix [nodes={draw=none, minimum size=4.5em},column sep={0em,between borders}, anchor=west] (result) at (2,0)
            {
                \node[fill=blue!30] {$f(a_1,a_2)$}; & \node[fill=orange!30] {$f(a_4,a_5)$}; \\
                \node[fill=red!30] {$b_3$}; & \node[fill=teal!30] {$f(b_2,b_3)$}; \\
                \node[fill=green!30] {$f(c_2,c_3)$}; & \node[fill=purple!30] {$f(c_0,c_1)$}; \\
                \node[fill=yellow!30] {$d_5$}; & \node[fill=violet!30] {$d_0$}; \\
            };
            \node at ($(result) - (0,4)$) {Result};
        \end{tikzpicture}
    \caption{Visualization of the \lstinline|select| operation on a \lstinline|RecordTensor|, where $a$, $b$, $c$, and $d$ are elements of the tensor at each time step, and $\Delta t = 1$ for simplicity. A $(T + 1) \times N$ record matrix is selected from with an $N \times D$ selector matrix, where $D$ is the number of samples to retrieve simultaneously, and each value is in the range $[0, T]$. The result matrix has the same shape as the latter, where $f$ is the interpolation function. Values are colored corresponding to the selection, where shading by multiple colors corresponds to multiple selections.}
    \label{fig:rt-select}
\end{figure}
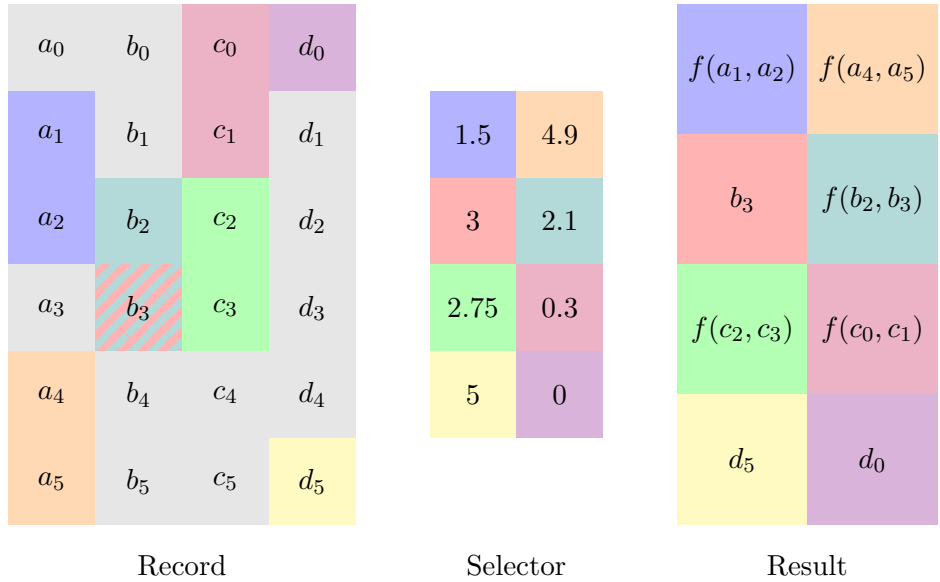

Prior observations can then be retrieved via the \lstinline|select| method by providing a tensor with values corresponding to the length of time before present (the most recent prior observation having occurred zero time ago) with the option to select multiple times for each recorded value. A new observation can be written in a similar way with the \lstinline|insert| method, or more commonly can be added as the newest observation with the \lstinline|push| method. Both \lstinline|select| and \lstinline|insert| employ the strategy pattern to support various functions for interpolating and extrapolating (which are defined by the protocols \lstinline|Interpolation| and \lstinline|Extrapolation| respectively). This is used to support continuous-time delays on discretely observed processes. A visualization of the \lstinline|select| operation can be seen in figure \ref{fig:rt-select}.

For instance, previous spikes can use \lstinline|interp_previous| to treat them as having occurred at any point during the time step. Processes described by exponential decay frequently arise in the simulation of SNNs in-part as the result of the solutions to the linear ODEs used to model the behavior of some of these systems. These can use \lstinline|interp_expdecay| or \lstinline|interp_expratedecay| to decay the value over the time since the last observation. Since an observation generally cannot occur between the discrete steps, this calculation is exact. Any custom behavior can be implemented by writing a function conforming to the \lstinline|Interpolation| or \lstinline|Extrapolation| protocols.

\subsubsection{Example: Delayed Linear Connections}
A linear connection (\lstinline|LinearDense|), with delay-learning enabled, is supported by delays with the shape $N_\text{out} \times N_\text{in}$, where $N_\text{out}$ is the number of outputs from the connection and $N_\text{in}$ is the number of inputs to the connection. Internally, a \lstinline|RecordTensor| is used to store prior observations, with a shape of

$$\left\lceil 1 + \frac{T}{\Delta t}\right\rceil \times B \times N_\text{in},$$

where $B$ is the batch size, $T$ is the maximum connection delay, and $\Delta t$ is the length of the time step of the simulation. All connections in Inferno have the \lstinline|selector| property which returns the learned delays (a tensor of zeros if the connection doesn't utilize delays), reshaped for use with \lstinline|RecordTensor| attributes. For \lstinline|LinearDense|, this has a shape of $1 \times N_\text{in} \times N_\text{out}$ and is used with the \lstinline|select| method to return the recorded values offset in time.

\begin{lstlisting}[caption={Create a dense linear connection with trainable delays.}, label=lst:delayed-linear-dense, language=Python]
import inferno.neural as snn

conn = snn.LinearDense(
    (28, 28),
    10,
    step_time=1.2,
    synapse=...,
    delay=6.0,
    batch_size=20,
    ...
)
\end{lstlisting}

Listing \ref{lst:delayed-linear-dense} constructs a dense linear connection with a simulation time step of $1.2\text{ms}$ and a maximum delay of $6\text{ms}$, which takes tensor(s) with a shape of $20 \times 28 \times 28$ as input and outputs a $20 \times 10$ tensor. Mapping this to the above, $B = 20$, $N_\text{in} = 784$, $N_\text{out} = 10$, $\Delta t = 1.2$, and $T = 6$. Inferno utilizes \lstinline|RecordTensor| internally and provides convenient properties for use with connections.

\begin{lstlisting}[caption={Access delayed state of a connection.}, label=lst:connection-delayed-access, language=Python]
...
conn.synapse.spike       # last spikes received
conn.synapse.current     # last currents received
conn.synspike            # last spikes used
conn.syncurrent          # last currents used
\end{lstlisting}

Listing \ref{lst:connection-delayed-access} continues from listing \ref{lst:delayed-linear-dense} and shows the properties that can be used to access the last received spikes and currents, as well as those that have been delay-shifted. Because the delays for inputs can be different across outputs, \lstinline|synspike| and \lstinline|syncurrent| have a shape of $B \times N_\text{in} \times N_\text{out}$, whereas \lstinline|synapse.spike| and \lstinline|synapse.current| have a shape of $B \times N_\text{in}$.

\subsubsection{Example: Monitors and Reducers}
In addition to the built-in records, Inferno provides a means of quickly recording derived properties. Listing \ref{lst:connection-delayed-access} continues from listing \ref{lst:delayed-linear-dense} and shows that in order to compute and record the spike trace (a value which is incremented on an action potential and decayed exponentially otherwise), a \lstinline|Monitor| with a corresponding \lstinline|Reducer| can be added.

\begin{lstlisting}[caption={Monitor the spike trace input for a connection.}, label=lst:monitor-spike-trace, language=Python]
...
from inferno.observe import (
    CumulativeTraceReducer,
    StateMonitor,
)

trace = StateMonitor(
    CumulativeTraceReducer(
        conn.dt,
        ...,
        target=True,
        duration=conn.delayedby,
        inclusive=False,
    ),
    attr="synapse.spike",
    module=conn,
    ...
)
\end{lstlisting}

On every forward pass into the connection, the state (in this case \lstinline|conn.synapse.spike|) will be recorded and decayed towards zero, then incremented for any spikes (inputs with a value of \lstinline|True|). \lstinline|CumulativeTraceReducer| also uses a \lstinline|RecordTensor|, and provides convenience methods to access that state after inputs have been presented to the connection. The argument \lstinline|inclusive| is passed into the \lstinline|RecordTensor| constructor and controls whether the sample at the duration should be stored.

\subsubsection{Example: {\tt RecordTensor} in a Custom Module}
Sometimes it may be necessary to directly interface with \lstinline|RecordTensor|. A \lstinline|RecordTensor| can be added as an attribute to a \lstinline|Module|. The class method \lstinline|create| simplifies this.

\begin{lstlisting}[caption={Creating a custom module with a \lstinline|RecordTensor|.}, label=lst:custom-recordtensor-module, language=Python]
from inferno import Module, RecordTensor
from inferno.functional import interp_linear


class EWMA(Module):

    def __init__(self, alpha, step_time, duration, inclusive):
        Module.__init__(self)
        self.alpha = min(0.0, max(1.0, float(alpha)))
        RecordTensor.create(
            self,
            "record",
            step_time,
            duration,
            ...,
            inclusive,
        )

    def __getitem__(self, time):
        return self.record.select(time, interp_linear, ...)

    def forward(self, value):
        prior = self.record.peek()
        if prior is None:
            self.record.push(value)
        else:
            self.record.push(
                self.alpha * value +
                (1 - self.alpha) * prior
            )
\end{lstlisting}

The code in listing \ref{lst:custom-recordtensor-module} defines an extension of \lstinline|Module| with a single \lstinline|RecordTensor| attribute named \lstinline|record| to track the exponentially weighted moving average (EWMA) of inputs. By default, its state will persist when saving and loading a model. When \lstinline|forward| is invoked, as it will be whenever the object is directly called (this pattern coming from \lstinline|nn.Module|), the most recent state will be retrieved. If no observations have been recorded, it will directly write the input to the record. Otherwise, it will compute the exponentially weighted moving average and write that value instead. When the object is accessed with square brackets, the \lstinline|RecordTensor| will retrieve the result from the previous time (or times) and interpolate linearly between observations when required.

\subsection{Shape Generalization for Pre/Post-Synaptic Training Regimes} \label{sec:generalized-shapes}
In order to facilitate the development and testing of new training regimes for spiking neural networks, Inferno provides generalizations that, once implemented for a specific kind of connection, will be compatible with all similar training regimes. The scope for these are training methods similar to spike-timing-dependent plasticity (STDP), specifically for regimes which compare the inputs to a connection and the outputs from a neuron model which receives input from the connection. In addition to two-factor methods, this also applies to three-factor methods \cite{fremaux_neuromodulated_2016} such as MSTDP (reward-modulated STDP) and MSTDPET (reward-modulated STDP with eligibility trace) \cite{florian_reinforcement_2007}.

For a connection supported by one or more parameters (e.g., weights), with a shape $P_1 \times P_2 \times \cdots \times P_n$, a product between the presynaptic and postsynaptic values has a shape

$$B \times P_1 \times P_2 \times \cdots \times P_n \times R,$$

where $B$ is the batch dimension and $R$ is a ``receptive field`` dimension. The latter of these corresponds to the outputs which are directly affected by a given element of the parameter tensor. All \lstinline|Connection| objects in Inferno have two methods corresponding to this, \lstinline|postsyn_receptive| and \lstinline|presyn_receptive|, in order to reshape outputs and inputs respectively. Inferno's implementation of STDP training regimes performs this efficiently using \lstinline|einops.einsum|, where \lstinline|einops| is a Python library for the implementation of tensor operations using Einstein summation notation \cite{rogozhnikov_einops_2022}. Specifically, the \lstinline|einsum| equation \lstinline|"b ... r, b ... r -> b ..."| will correctly take the necessary product with or without learned delays.

\subsubsection{Example: Linear Connection Updates} \label{sec:lin-conn}
A linear connection is supported by weights with the shape $N_\text{out} \times N_\text{in}$, where $N_\text{out}$ is the number of outputs from the connection and $N_\text{in}$ is the number of inputs to the connection. Each weight corresponds to the relation between a single input and output. Therefore, the postsynaptic values are shaped like $B \times N_\text{out} \times 1 \times 1$ and presynaptic values are reshaped like $B \times 1 \times N_\text{in} \times 1$ if not delayed, and $B \times N_\text{out} \times N_\text{in} \times 1$ otherwise. Although both cases are represented as the same einsum operation, when there is no learned delay, the operation is equivalent to batched matrix multiplication. When there is a learned delay, the operation is instead equivalent to the Hadamard product.

\subsubsection{Example: Two-Dimensional Convolutional Connection Updates} \label{sec:conv-conn}
A two-dimension convolutional connection is supported by weights---typically called the \emph{kernel}---with the shape $F \times C \times k_H \times k_W$. Here $F$ is the number of filters (each responsible for an output channel), $C$ is the number of input channels, and $k_H$ and $k_W$ are the height and width of the convolutional kernel respectively. For each dimension $d$ along which the convolution occurs, the number of outputs is computed as

$$L_d = \left\lfloor \frac{d + 2 p_d - l_d (k_d - 1) - 1}{s_d} \right\rfloor + 1,$$

where $d$ is the size of dimension, $p_d$ is the padding added to each side, $l_d$ is the dilation of the convolution \cite{yu_multi-scale_2016}, $k_d$ is the kernel size, and $s_d$ is the stride of the convolution. The number of outputs for a convolution is therefore $L = \prod_d L_d$. The efficient and parallelizable computation of convolutions is performed by unfolding (also called unrolling, and in the two-dimensional case, im2col) the convolution blocks and performing matrix-matrix multiplication with the kernel as given in \cite{chellapilla_high_2006}. In the two-dimensional case, this results in a tensor with a shape of

$$B \times (C \cdot k_H \cdot k_W) \times L,$$

where $C \cdot k_H \cdot k_W$ is the size of each convolutional block and $L$ is the number of blocks (i.e. the number of inputs to which each parameter was applied). Inferno reshapes postsynaptic values into the form $B \times F \times 1 \times 1 \times 1 \times L$, but for presynaptic values, the reshaping depends on whether the connection has trainable delays. The resulting shape is $B \times 1 \times C \times k_H \times k_W \times L$ when there is no delay, otherwise it is $B \times F \times C \times k_H \times k_W \times L$. As with the case given in section \ref{sec:lin-conn}, an einsum operation will collapse these to a tensor shaped like the parameter itself, with an additional batch dimension.

\subsubsection{Implications}
A principal design objective of Inferno was to facilitate the rapid development of new techniques for SNNs with applications in machine learning. The \lstinline|Connection| classes provide high-level abstractions that enable this by disentangling the implementation of any specific \lstinline|Connection| from training methods. When implementing a new training method with Inferno, building it around the \lstinline|postsyn_receptive| and \lstinline|presyn_receptive| methods provided by \lstinline|Connection| means it can automatically work with \emph{all} \lstinline|Connection| classes.

BindsNET requires each training method to be implemented separately for \emph{each} class. Likewise, any new type of connection requires reimplementing each training method for it. The result is a slow, cumbersome, and error-prone process that hinders the end-user from being able to develop and test new techniques at scale.

Brian 2 lacks the high-level abstractions provided by Inferno and BindsNET. There are no built-in connections to use with newly developed training methods or vice versa. Code can be more shareable as the logic is often written in a Python-like domain-specific language. The lack of high-level constructs often used in deep learning slows development and makes it more error-prone. Brian 2 excels at what it was designed for, but its objectives diverge from those of Inferno, and in this case exacerbates these issues. It is extremely customizable, but the lack of structured components makes both code reuse and the organization of larger projects difficult.

\subsection{Generalized Parameter-Dependent Updates} \label{sec:generalized-updates}
When training a model using STDP or a variant thereof, the magnitude of each connection's weight updates may be scaled according to a weight dependence rule, such as power-law weight dependence \cite{gutig_learning_2003}. This scales the magnitudes of potentiative weight updates $\Delta w_+$ and depressive weight updates $\Delta w_-$ as

\begin{align*}
    S(\Delta w_+) &= (w_\text{max} - w)^{\mu_+} \Delta w_+, \\
    S(\Delta w_-) &= (w - w_\text{min})^{\mu_-} \Delta w_-,
\end{align*}

where $w_\text{max}$ and $w_\text{min}$ are the upper and lower bounds for connection weights, $\mu_+$ and $\mu_-$ are the weight dependence powers for the upper and lower bounds, and $w$ is the value of the connection weights.

Rather than tying this method of parameter-dependent updates to a given method (such as any specific implementation of STDP) or to any specific parameter (such as connection weights), Inferno allows for this to be easily generalized.
\begin{itemize}
    \item \lstinline|Connection| objects inherit from \lstinline|Updatable|, which allows for an \lstinline|Updater| to be attached which is used to apply updates to the trainable parameters of a \lstinline|Connection|. The \lstinline|Updater| will create an \lstinline|Accumulator| for each trainable parameter.
    \item The \lstinline|Accumulator| can then be given either a potentiative update, depressive update, or tuple of the two. If an \lstinline|Accumulator| is given multiple of either type of update, they will be reduced together (by taking the mean, sum, etc.) and parameter dependence will then be applied before updating the parameter. This is in-line with Inferno's emphasis on code reusability, especially with regard to training methods, as previously discussed in section \ref{sec:generalized-shapes}.
    \item Inferno provides the \lstinline|HalfBounding| and \lstinline|FullBounding| protocols as well as a number of implemented functions, including for power-law dependence, which can be passed into the \lstinline|Accumulator| for a given \lstinline|Connection| parameter.
\end{itemize}

\subsubsection{Example: Bounding and Updating Connection Weights}
The following code examples will demonstrate how these abstractions can be used to provide connection parameter bounding, and how to apply parameter updates.

\begin{lstlisting}[caption={Create an updater and add parameter bounding.}, label=lst:add-param-dependence, language=Python]
import inferno.functional as inff
import inferno.neural as snn

conn = ...
conn.updater = conn.defaultupdater

conn.updater.weight.lowerbound(inff.bound_lower_sharp, 0.0)
conn.updater.weight.upperbound(inff.bound_upper_power, 1.0, power=0.5)
\end{lstlisting}

Letting \lstinline|conn| be a \lstinline|Connection| object, the rest of the code in listing \ref{lst:add-param-dependence} adds the default \lstinline|Updater| for that \lstinline|Connection| to \lstinline|conn|. The trainable parameters are set as properties on the \lstinline|Updater| dynamically at runtime. The \lstinline|Accumulator| for \lstinline|conn.weight| has power-law dependence applied to its upper bound, with $w_\text{max} = 1$ and $\mu_+ = 1/2$. The lower bound is instead controlled with sharp bounding (also called hard weight dependence) where updates are only applied if the parameter is not outside the permitted bounds, here with $w_\text{min} = 0$ \cite{gerstner_neuronal_1996}.

Once the potentiative and depressive components of the \lstinline|Updater| are computed, they can be added to the \lstinline|Accumulator|, which will automatically apply the specified bounding method on its update.

\begin{lstlisting}[caption={Add and apply connection updates.}, label=lst:apply-acc-updates, language=Python]
...

update_pos = ...
update_neg = ...

conn.updater.weight.pos = update_pos
conn.updater.weight.neg = update_neg

conn.update()
\end{lstlisting}

Listing \ref{lst:apply-acc-updates} continues from listing \ref{lst:add-param-dependence}. In the latter, we let \lstinline|update_pos| and \lstinline|update_neg| be tensors of nonnegative values represented the potentiative and depressive parameter updates respectively. These updates are then appended to the list of positive and negative updates for \lstinline|conn.weight|. When \lstinline|conn.update| is called, the positive and negative updates are reduced separately (applicable if multiple updates are appended), and then applied using the previously set parameter bounding functions.

\section{Implementing Delay Learning}
In a recently published paper \cite{nadafian_bioplausible_2024}, Nadafian and Ganjtabesh introduce how STDP can be applied to the task of learning both weights and delays. Their methods for each of these tasks are similar and involve adjusting the time difference between presynaptic and postsynaptic spikes using the values of the learned delays. In this section, we will show how the delay learning component of this is implemented by Inferno. Henceforth, we will refer to this as ``delay-adjusted STDP for delays". The update rule is as follows:

$$
D(t + \Delta t) - D(t) =
\begin{cases}
    B_- \exp\left(-\frac{\lvert t_\Delta(t) \rvert}{\tau_-} \right) &t_\Delta(t) \geq 0 \\
    B_+ \exp\left(-\frac{\lvert t_\Delta(t) \rvert}{\tau_+} \right) &t_\Delta(t) < 0
\end{cases}
$$

$$t_\Delta(t) = t^f_\text{post} - t^f_\text{pre} - D(t).$$

The hyperparameters $B_-$ and $B_+$ control the magnitudes of the updates for causal (pre-then-post) and anti-causal (post-then-pre) spike pairs respectively. The updates are Hebbian when $B_- < 0$ and $B_+ > 0$. The time constants of exponential decay for these updates are $\tau_-$ and $\tau_+$, respectively. The delay-adjusted spike time difference $t_\Delta$ for each connected neuron pair is calculated from the time of the last postsynaptic and presynaptic spikes, $t^f_\text{post}$ and $t^f_\text{pre}$ respectively, and the delay between them $D(t)$ at time $t$.

Subclasses of Inferno's \lstinline|IndependentCellTrainer| divide the computation of updates across two methods, \lstinline|register_cell| and \lstinline|forward|. The former of these is responsible for hooking any required \lstinline|Monitor| objects to the \lstinline|Cell| being trained. The latter is responsible for using these intermediate values to compute the final update term and passing it into the relevant \lstinline|Accumulator|.

\begin{lstlisting}[caption={The \lstinline|register_cell| method for delay-adjusted STDP for delays.}, label=lst:dadj-stdp-delay-regcell, language=Python]
def register_cell(self, name, cell, /, **kwargs):
    cell, state = self.add_cell(
        name, cell, self._build_cell_state(**kwargs), ["delay"]
    )
    monitor_kwargs = {
        "as_prehook": False,
        "train_update": True,
        "eval_update": False,
        "prepend": True,
    }
    self.add_monitor(
        name, "spike_post", "neuron.spike",
        StateMonitor.partialconstructor(
            reducer=EventReducer(
                cell.connection.dt,
                lambda x: x.bool(),
                initial="nan",
                inplace=state.inplace
            ),
            **monitor_kwargs,
        ),
        False,
        dt=cell.connection.dt,
    )
    self.add_monitor(
        name, "spike_pre", "synapse.spike",
        StateMonitor.partialconstructor(
            reducer=EventReducer(
                cell.connection.dt,
                lambda x: x.bool(),
                initial="nan",
                inplace=state.inplace
            ),
            **monitor_kwargs,
        ),
        False,
        dt=cell.connection.dt,
    )
    return self.get_unit(name)
\end{lstlisting}

The code in listing \ref{lst:dadj-stdp-delay-regcell} creates two \lstinline|Monitor| objects, one targeting the spikes produced by the \lstinline|Neuron| object and the other the spikes received by the \lstinline|Synapse| object. The latter \emph{excludes} the offset from learned delays as this is factored into the $t_\Delta(t)$ term separately. Each \lstinline|Monitor| uses an \lstinline|EventReducer|, a \lstinline|Reducer| class that tracks the duration since the last event (given here by casting the observation itself into a boolean datatype). The parameter \lstinline|initial| controls the value before any observation as either \lstinline|NaN|, $+\infty$, or $0$. Here \lstinline|NaN| is used so that updates before both a postsynaptic and presynaptic occur can be filtered out easily.

\begin{lstlisting}[caption={The \lstinline|forward| method for delay-adjusted STDP for delays.}, label=lst:dadj-stdp-delay-forward, language=Python]
def forward(self) -> None:
    for cell, state, monitors in self:
        if not cell.training or not self.training or not cell.updater:
            continue

        t_post = cell.connection.postsyn_receptive(
            monitors["spike_post"].peek()
        )
        t_pre = cell.connection.presyn_receptive(
            monitors["spike_pre"].peek()
        )

        t_delta = t_pre - t_post - cell.connection.delay.unsqueeze(-1)
        t_delta_abs = t_delta.abs()

        mask = (t_delta >= 0).to(dtype=t_delta_abs.dtype)
        dneg = state.batchreduce(
            (
                torch.exp(t_delta_abs / (-state.tc_neg))
                * (abs(state.lr_neg) * mask)
            ).nansum(-1),
            0,
        )

        mask = (t_delta < 0).to(dtype=t_delta_abs.dtype)
        dpos = state.batchreduce(
            (
                torch.exp(t_delta_abs / (-state.tc_pos))
                * (abs(state.lr_pos) * mask)
            ).nansum(-1),
            0,
        )

        match (state.lr_neg < 0, state.lr_pos < 0):
            case (True, True):
                cell.updater.delay = (None, dpos + dneg)
            case (True, False):
                cell.updater.delay = (dpos, dneg)
            case (False, True):
                cell.updater.delay = (dneg, dpos)
            case (False, False):
                cell.updater.delay = (dpos + dneg, None)
\end{lstlisting}

The code in listing \ref{lst:dadj-stdp-delay-forward} takes the tensor of times since the last postsynaptic spikes and the tensor of times since the last presynaptic spikes and applies the reshaping operations \lstinline|postsyn_receptive| and \lstinline|presyn_receptive| respectively to enable this code to work for any \lstinline|Connection|. Calculating the $t_\Delta(t)$ term precludes using \lstinline|einops.einsum| as is used in the implementations of pair-based and triplet STDP. Using \lstinline|nansum| along the receptive dimension filters out any invalid updates. Assignment of the positive and negative update components to the \lstinline|cell.updater.delay| property allows for parameter dependence to be applied when the update itself is applied.

Listings \ref{lst:dadj-stdp-delay-regcell} and \ref{lst:dadj-stdp-delay-forward} show how it is possible to go from the formula for a newly developed training method to having an implementation of it that works across network architectures rapidly, and in a manner that leverages the generalizability provided by Inferno.

\section{Benchmarking}
For testing performance, we developed two kinds of benchmarks, one representing an inference task and the other a training task. The intention of these benchmarks is to represent a minimal, end-to-end pipeline. For these comparisons, we tested Inferno against Brian 2 and BindsNET. The former was chosen for its widespread use in the simulation of SNNs. The latter was chosen as it is the largest in active development with a specific focus on machine learning applications. Brian2CUDA was used for compiling Brian 2 to CUDA kernels when benchmarking on GPU.

Inferno and BindsNET directly execute Python code whereas Brian 2 and Brian2CUDA require a lengthy code generation and compilation stage. The time taken for that stage is excluded from all benchmark measurements. Execution times for Inferno and BindsNET were measured using Python's built-in timing functions after ensuring that all CUDA kernels finished executing. Execution times for Brian 2 and Brian2CUDA were measured using Brian 2's profiling tool.

All benchmarks were run on a Windows 10 desktop with an AMD Ryzen 5950X, 64 GB of system memory, and an NVIDIA RTX 3090. Brian 2 and Brian2CUDA were run in Ubuntu 22.04 LTS via WSL2. Benchmarks at each scale were consecutively run 15 times. The first 5 runs were treated as ``warm-up'' and the results discarded while the remaining 10 results were averaged. All models were run with 32-bit precision floating point values.

\subsection{PDL Benchmark} \label{sec:pdl-benchmark}
Based on the benchmark from the BindsNET paper \cite{hazan_bindsnet_2018}, this test is designed to emulate a complete pipeline for model inference. It consists of generating $N$ spike trains using a homogeneous Poisson process and transforming these discrete spikes into currents with the delta synapse model. At each time step, these $N$ currents are mapped to $N$ LIF neurons. This mapping is performed with a dense linear connection, parameterized by $N^2$ weights.

We attempted to model this in the most similar way possible across libraries. BindsNET does not support synaptic modeling, although its implementation of LIF neurons assumes the charge (in picocoulombs) has the same value as the membrane time constant (in milliseconds) if using the delta synapse model. The implementation of homogeneous Poisson spike generation also varies. Inferno generates spike times by sampling from an exponential distribution and taking the cumulative sum of these intervals. Brian 2 uses a common approximation by sampling from a uniform distribution and applying a threshold criterion to determine if a spike was generated. BindsNET strictly does not have a way to generate a Poisson spike train---it uses a similar approach to Inferno but determines spike times by taking the cumulative sum of values sampled from a Poisson distribution\footnote{We believe this to be an error in the library at the time of writing. This is supported by the fact that Inferno implements all three encoding schemes and only the BindsNET-like implementation (called named \lstinline|PoissonIntervalEncoder| to disambiguate) yields a significantly different spike train pattern. Specifically, the interspike intervals are far more evenly distributed.}.

The expected frequency of spikes for each spike train was sampled from $\mathcal{U} (0, 250) \text{ Hz}$. Each of the weights in the $N \times N$ weight matrix was sampled from $\mathcal{U} (0, 1)$. Each LIF neuron has a rest voltage of $-60 \text{mV}$, a reset voltage of $-65 \text{mV}$, a voltage threshold of $-50 \text{mV}$, a decay time constant of $20 \text{ms}$, a membrane resistance of $1 \text{M}\Omega$, and an absolute refractory period of $3 \text{ms}$. The simulation is run for $1000 \text{ms}$ of simulated time, with a step length of $1 \text{ms}$.

\begin{figure}[ht!]
    \centering
    \includegraphics[width=\linewidth]{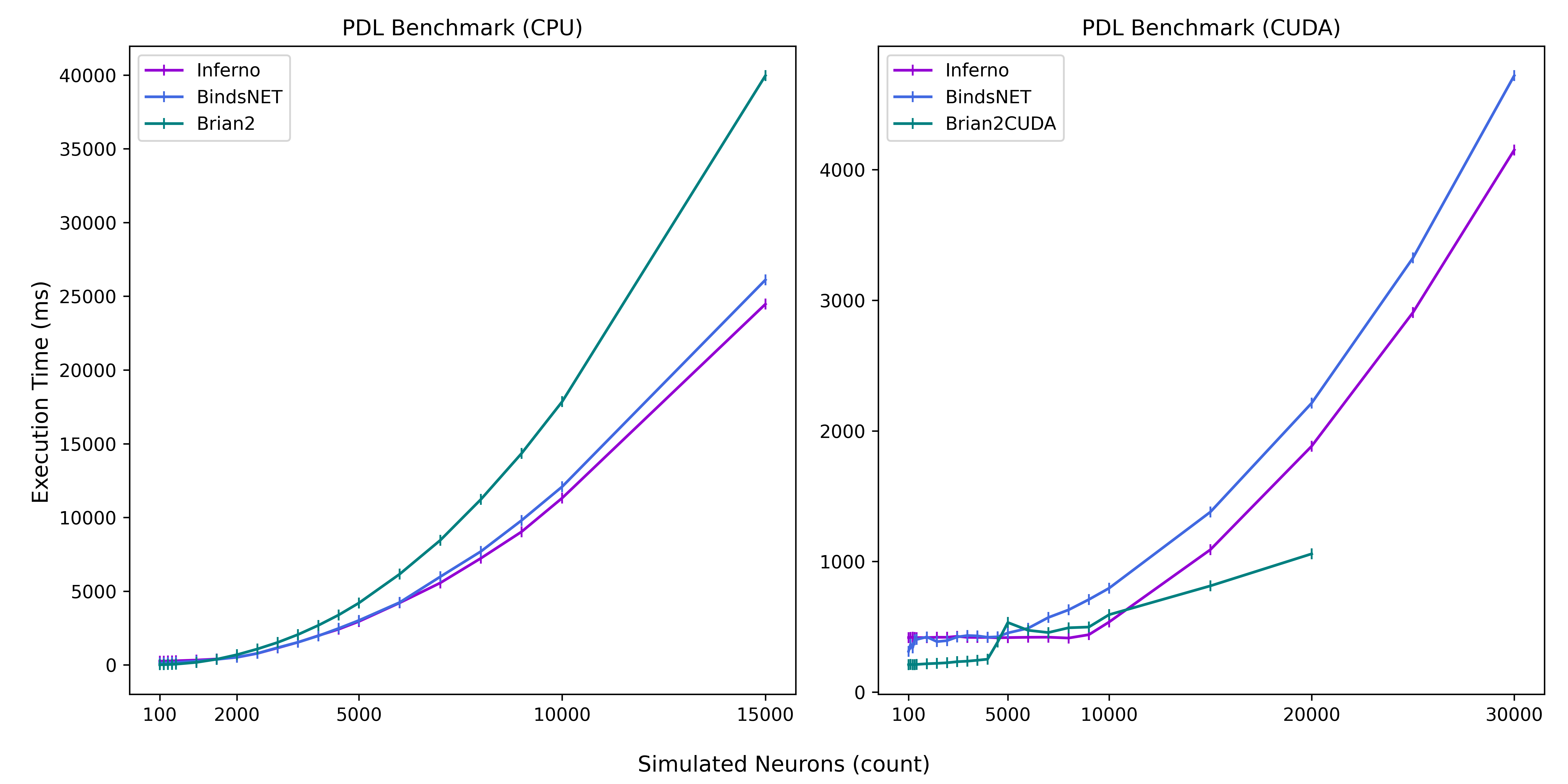}
    \caption{Execution times for the Poisson--Dense--Linear (PDL)---simulated for 1000 steps with a step length of $1 \text{ms}$. The benchmark for Brian2CUDA failed to complete for tests with more than 20,000 neurons.}
    \label{fig:pdl-results}
\end{figure}

As shown in figure \ref{fig:pdl-results}, when running on CPU, Inferno and BindsNET are significantly more performant than Brian 2, with Inferno maintaining a slight edge. For GPU execution the results are less clear. Brian2CUDA was more performant when simulating fewer than 5000 neurons (25 million weights) and more than 10,000 neurons (100 million weights), but was slower than both Inferno and BindsNET in-between. Execution of the benchmark with more than 20,000 neurons failed with Brian2CUDA. We were unable to diagnose the reason for this failure, but do not believe it to have been caused by a lack of hardware resources. For all but the smallest tested groups of neurons, Inferno was as fast as or faster than BindsNET in this test.

\subsection{STDP + PDL Benchmark}
In order to measure the training performance of Inferno, we extend the benchmark described in section \ref{sec:pdl-benchmark}. After each inference step, the weights are updated using STDP and are clipped to fall within the range $[0, 1]$. The magnitudes of the learning rates for potentiative and depressive learning rates were set to $10^{-3}$, signed so that the updates are Hebbian. The time constants for presynaptic and postsynaptic spike traces were both set to $20 \text{ms}$.

\begin{figure}[ht!]
    \centering
    \includegraphics[width=\linewidth]{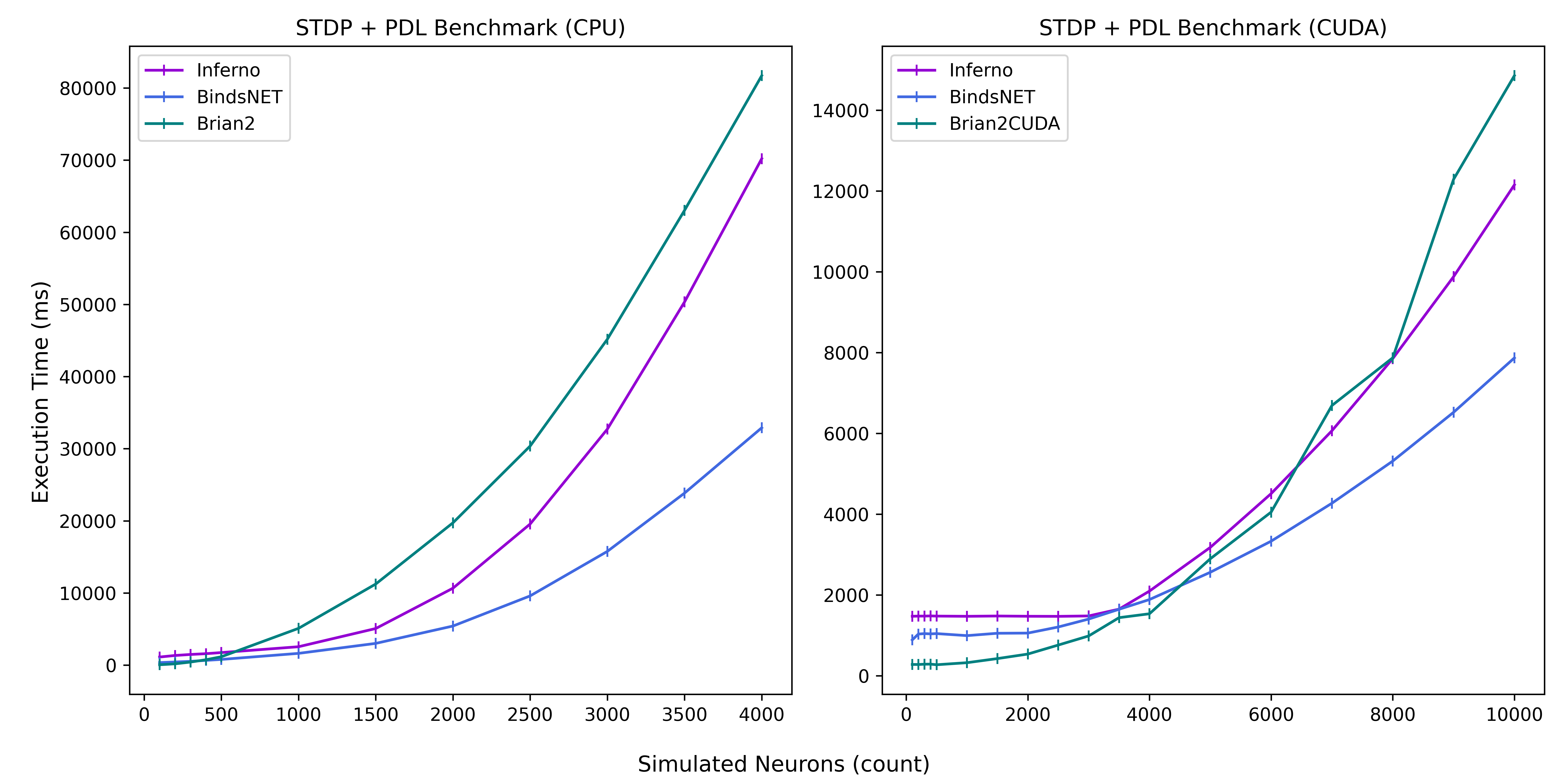}
    \caption{Execution times for the STDP + PDL benchmark---simulated for 1000 steps with a step length of $1 \text{ms}$.}
    \label{fig:stdp-results}
\end{figure}

As shown in figure \ref{fig:stdp-results}, when executing on CPU the performance of Inferno falls between that of Brian 2 and BindsNET for neuron counts over 500. On GPU, Inferno is slower with 6000 or fewer neurons, only gaining a slight advantage over Brian2CUDA for larger neuron counts. The curves appear to follow similar trajectories. BindsNET is consistently the fastest overall.

Inferno sacrifices some performance for greater flexibility. BindsNET requires a distinct implementation of each training method for each type of connection (linear, two-dimensional convolutional, etc.). With Inferno, developers can write a new training method only once and have it apply to all \lstinline|Connection| types. Likewise, if they write a new \lstinline|Connection|, implementing its required methods will ensure it is compatible with every \lstinline|Trainer|, not requiring any re-implementation. For some training methods, including STDP, Inferno can also train a \lstinline|Connection| in a ``delay-aware'' manner such that the connection delays can be updated alongside connection weights.

\section{Discussion}
Inferno was designed to bridge between research into novel SNN models and their application to real-world machine learning problems. Built atop PyTorch and following its conventions, Inferno not only leverages its high performance tensor operations on both CPU and GPU, but also makes developing with Inferno familiar to anyone who has previously used PyTorch. This both eliminates the lengthy code generation stage required of Brian 2 on GPU and preserves the interactivity normally only present when executing on CPU. It also means that improvements made to PyTorch will apply to Inferno, such as execution with Metal Performance Shaders on supported Apple Silicon devices, a feature released in early 2023 with PyTorch 2.0.

As presented, Inferno sacrifices some generalizability over Brian 2 but in exchange provides an interface that is far more approachable to machine learning researchers less demanding on details of the neuroscience behind SNNs. Inferno is similar in spirit to BindsNET, but maintains a more flexible and extensible approach, facilitating the development of new models and training methods. It also adds support for learned synaptic delays, an actively researched area of SNNs. Inferno is well positioned to provide a common framework for the development of SNN models, and to make SNNs accessible as a tool for machine learning tasks. Inferno's source code is hosted on GitHub at \url{https://github.com/mdominijanni/inferno} and documentation is available at \url{https://docs.inferno-ai.dev/}.

\printbibliography

\appendix

\section{Inferno Library}
\subsection{Package Layout}
\begin{description}[font=\ttfamily\bfseries]
    \item[inferno] The common infrastructure used throughout various submodules and various functions for calculations and tensor-creation which may be used either within Inferno or may be helpful for end-users.
    \item[inferno.extra] A work-in-progress module containing any ``dummy'' components which may be useful when attempting to generate visualizations or diagnose issues.
    \item[inferno.functional] The protocols and various implementations for parameter bounding, interpolation, extrapolation, and dimensionality reduction.
    \item[inferno.learn] The components needed for training spiking neural networks, as well as components which may be used for specific inference tasks (e.g. classification).
    \item[inferno.neural] The basic components for spiking neural networks, the infrastructure used for connecting them into a network and for supporting generalized parameter updates, and encoding non-spiking data into spike trains.
    \item[inferno.neural.functional] The functional implementation of various components used by different models as a way to generalize and share functionality, also useful when implementing new classes that represent neural components.
    \item[inferno.observe] The infrastructure and components for monitoring the internal states of components.
    \item[inferno.stats] A work-in-progress module containing PyTorch-based implementations of various probability distributions.
\end{description}

\subsection{Infrastructure Classes}
Below includes the basic infrastructure classes included in the top-level \lstinline|inferno| module. These are used extensively throughout the entire codebase.

\begin{description}[font=\ttfamily\bfseries]
    \item[Module] An extension of PyTorch's \lstinline|nn.Module| which includes a convenience function to persist non-tensor state when checkpointing a model, as well as an enhancement which enables the use of Python descriptors (including properties) which entail the assignment of submodules (nested \lstinline|nn.Module| objects).
    \item[ShapedTensor] Uses reflection to register a tensor to a \lstinline|Module| with additional runtime checks placed on the allowable shape of a tensor through dimensional constraints, as well as the ability to reconstrain (automatically reshape) a tensor which no longer meets those constraints.
    \item[VirtualTensor] Uses reflection to allow for a tensor-property to be added to a \lstinline|Module| which preserves the datatype and device set using \lstinline|nn.Module.to|. Useful if a property should return a floating point tensor but is derived from a boolean tensor (such as one storing spikes).
    \item[RecordTensor] Uses reflection to add a \lstinline|ShapedTensor| to a \lstinline|Module| with additional support for temporal indexing with continuous values, including interpolation between recorded observations, as well as assignment in the same way with extrapolation.
    \item[Hook] Managed calls of PyTorch forward hooks. It includes filtering of activation based on the state of \lstinline|nn.Module.training|.
    \item[ContextualHook] Variant of \lstinline|Hook| with activation-calls based on weak references to itself, allowing for either using a method or callable attributes without creating cyclic references, used to avoid garbage collection issues.
    \item[StateHook] Limited version of \lstinline|ContextualHook| which can only act on module state (not on the arguments or return value of an \lstinline|nn.Module| call), but which can be triggered manually as well.
\end{description}

\section{End-to-End Example}
Given in listing \ref{lst:end-to-end-eg} is a complete example of using Inferno to build a model, training that model to perform a classification task, and evaluating its performance. The example is based on the ``quickstart`` example provided by PyTorch \cite{paszke_pytorch_2019}. The model implemented is based on the one described in \cite{diehl_unsupervised_2015}. This example performs classification on the MNIST dataset \cite{lecun_gradient-based_1998}. Default values are provided, but with slight modification they can be controlled by some configuration.

\begin{lstlisting}[caption={End-to-End Code Example}, label=lst:end-to-end-eg, language=Python]
import inferno
import torch
from inferno import functional, neural, learn
from torch.utils.data import DataLoader
from torchvision.datasets import MNIST
from torchvision.transforms.v2 import Compose, ToImage, ToDtype, Lambda


# runtime values
shape = (10, 10)
step_time = 1.0
batch_size = 20

if torch.cuda.is_available():
    device = "cuda"
elif torch.backends.mps.is_available():
    device = "mps"
else:
    device = "cpu"

print(f"using device: {device}\n")


# retrieve and load the data
train_set = MNIST(
    root="data",
    train=True,
    transform=Compose(
        [
            ToImage(),
            ToDtype(torch.float32, scale=True),
            Lambda(lambda x: x.squeeze(0)),
        ],
    ),
    download=True,
)

test_set = MNIST(
    root="data",
    train=False,
    ...  # rest as above
)

train_data = DataLoader(train_set, batch_size=batch_size)
test_data = DataLoader(test_set, batch_size=batch_size)


# construct the neurons, most values follow bindsnet
exc = neural.ALIF(
    shape,
    step_time,
    rest_v=-65.0,
    reset_v=-60.0,
    thresh_eq_v=-52.0,
    refrac_t=5.0,
    tc_membrane=100.0,
    tc_adaptation=1e7,
    spike_increment=0.05,
    resistance=1.0,
    batch_size=batch_size,
)

inh = neural.LIF(
    shape,
    step_time,
    rest_v=-60.0,
    reset_v=-45.0,
    thresh_v=-40.0,
    refrac_t=2.0,
    time_constant=75.0,
    resistance=1.0,
    batch_size=batch_size,
)


# construct the connections, most values follow bindsnet
enc2exc = neural.LinearDense(
    (28, 28),
    shape,
    step_time,
    synapse=neural.DeltaCurrent.partialconstructor(100.0),
    weight_init=lambda x: inferno.rescale(inferno.uniform(x), 0.0, 0.3),
    batch_size=batch_size,
)

inh2exc = neural.LinearLateral(
    shape,
    step_time,
    synapse=neural.DeltaCurrent.partialconstructor(100.0),
    weight_init=lambda x: inferno.full(x, -180.0),
    batch_size=batch_size,
)

exc2inh = neural.LinearDirect(
    shape,
    step_time,
    synapse=neural.DeltaCurrent.partialconstructor(75.0),
    weight_init=lambda x: inferno.full(x, 22.5),
    batch_size=batch_size,
)

# build the model class
class Model(inferno.Module):

    def __init__(self, exc, inh, enc2exc, inh2exc, exc2inh):

        # call superclass constructor
        inferno.Module.__init__(self)

        # construct the layers
        self.feedfwd = neural.Serial(enc2exc, exc)
        self.inhibit = neural.Serial(inh2exc, exc)
        self.trigger = neural.Serial(exc2inh, inh)

    def forward(self, inputs, trainer=None):
        # clears the model state
        def clear(m):
            if isinstance(m, neural.Neuron | neural.Connection):
                m.clear()

        # compute for each time step
        def step(x):

            # inference
            res = self.feedfwd(x)
            _ = self.inhibit(self.trigger(res))

            # training
            if self.training and trainer:
                trainer()
                self.feedfwd.connection.update()

            return res

        res = torch.stack([step(x) for x in inputs], dim=0)
        self.apply(clear)

        return res

model = Model(exc, inh, enc2exc, inh2exc, exc2inh)
model.to(device=device)

# add weight normalization
norm_hook = neural.Normalization(
    model,
    "feedfwd.connection.weight",
    order=1,
    scale=78.4,
    dim=-1,
    eval_update=False,
)
norm_hook.register()
norm_hook()


# create the trainer and connect it
trainer = learn.STDP(
    step_time=step_time,
    lr_post=5e-4,
    lr_pre=-5e-6,
    tc_post=30.0,
    tc_pre=30.0,
)

updater = model.feedfwd.connection.defaultupdater()
model.feedfwd.connection.updater = updater

trainer.register_cell("feedfwd", model.feedfwd.cell)
trainer.to(device=device)


# add weight bounding
clamp_hook = neural.Clamping(
    updater,
    "parent.weight",
    min=0.0,
)
clamp_hook.register()

updater.weight.upperbound(
    functional.bound_upper_multiplicative,
    1.0,
)


# build the classifier
classifier = learn.MaxRateClassifier(
    shape,
    num_classes=10,
    decay=1e-6,
)
classifier.to(device=device)


# build the encoder
encoder = neural.HomogeneousPoissonEncoder(
    250,
    step_time,
    frequency=128.0,
)



# create and run the training/testing loop
classify_interval = 10

def train(data, encoder, model, trainer, classifier):
    size = len(data.dataset)
    rates, labels = [], []
    correct, current = 0, 0

    for batch, (X, y) in enumerate(data, start=1):
        X, y = X.to(device=device), y.to(device=device)

        rates.append(model(encoder(X), trainer).float().mean(dim=0))
        labels.append(y)

        if batch % classify_interval == 0:
            rates = torch.cat(rates, dim=0)
            labels = torch.cat(labels, dim=0)

            pred = classifier(rates, labels)
            nc = (pred == labels).sum().item()
            correct += nc
            current += rates.size(0)

            print(f"acc: {(nc / rates.size(0)):.4f} "
                  f"[{current:>5d}/{size:>5d}]")
            rates, labels = [], []

    print(f"Training Accuracy:\n    {(correct / size):.4f}")

def test(data, encoder, model, classifier):
    size = len(data.dataset)
    correct, current = 0, 0

    for batch, (X, y) in enumerate(data, start=1):
        X, y = X.to(device=device), y.to(device=device)

        rates = model(encoder(X), None).float().mean(dim=0)
        pred = classifier(rates, None)

        correct += (pred == y).sum().item()
        current += rates.size(0)


    print(f"Testing Accuracy:\n    {(correct / size):.4f}\n")

epochs = 1

with torch.no_grad():
    for epoch in range(epochs):
        print(f"Epoch {epoch + 1}\n-------------------------")
        model.train(), trainer.train()
        train(train_data, encoder, model, trainer, classifier)
        model.eval(), trainer.eval()
        test(test_data, encoder, model, classifier)
    print("Completed")
\end{lstlisting}

\end{document}